\documentclass[runningheads,a4paper]{svmult}
\usepackage{type1cm}        % activate if the above 3 fonts are
\usepackage{makeidx}         % allows index generation
\usepackage{graphicx}        % standard LaTeX graphics tool
\usepackage{subfigure}
\usepackage{multicol}        % used for the two-column index

\usepackage{tabularx,booktabs}
\usepackage[bottom]{footmisc}% places footnotes at page bottom

\usepackage{newtxtext}       % 
\usepackage[varvw]{newtxmath}       % selects Times Roman as basic font

\usepackage{hyperref}
\usepackage[xcolor={divpdf,grey},authormarkup=none]{changes} 
\usepackage{academicons}
\usepackage{svg}

\newcommand{\orcid}[1]{\href{https://orcid.org/#1}{\includegraphics[width=0.03\textwidth]{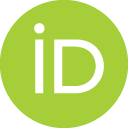}}}
\makeindex             % used for the subject index
\begin{document}

% \title*{Wykorzystanie Systemów Wizyjnych i Digital Twin do Utrzymania Czystości w Przestrzeniach Publicznych}
\title*{Utilisation of Vision Systems and Digital Twin for Maintaining Cleanliness in Public Spaces}
\titlerunning{Utilisation of Vision Systems and Digital Twin for Maintaining Cleanliness}
% Use \titlerunning{Short Title} for an abbreviated version of
% your contribution title if the original one is too long
\author{
Mateusz Wasala \orcid{0000-0002-8631-8428}, %\and
Krzysztof Blachut \orcid{0000-0002-1071-335X}, %\and 
Hubert Szolc \orcid{0000-0003-3018-5731}, %\and 
Marcin Kowalczyk \orcid{0000-0002-4257-8877}, %\and
Michal Danilowicz \orcid{0000-0003-3018-5731}, %\and
Tomasz Kryjak \orcid{0000-0001-6798-4444} 
} 

\authorrunning{M. Wasala et al.}
% First names are abbreviated in the running head.
% If there are more than two authors, 'et al.' is used.
%
\institute{Embedded Vision Systems Group, \\ Department of Automatic Control and Robotics, \\ AGH University of Krakow, Poland \\
\email{{wasala, kblachut, szolc, kowalczyk, danilowi, tomasz.kryjak}@agh.edu.pl} 
}
%
% Use the package "url.sty" to avoid
% problems with special characters
% used in your e-mail or web address
%
\maketitle

\vspace{-60pt}
\abstract{Nowadays, the increasing demand for maintaining high cleanliness standards in public spaces results in the search for innovative solutions.  The deployment of CCTV systems equipped with modern cameras and software enables not only real-time monitoring of the cleanliness status but also automatic detection of impurities and optimisation of cleaning schedules. The Digital Twin technology allows for the creation of a~virtual model of the space, facilitating the simulation, training, and testing of cleanliness management strategies before implementation in the real world.
In this paper, we present the utilisation of advanced vision surveillance systems and the Digital Twin technology in cleanliness management, using a~railway station as an example. The Digital Twin was created based on an actual 3D model in the Nvidia Omniverse Isaac Sim simulator. A~litter detector, bin occupancy level detector, stain segmentation, and a~human detector (including the cleaning crew) along with their movement analysis were implemented. A~preliminary assessment was conducted, and potential modifications for further enhancement and future development of the system were identified.}

\keywords{vision systems, image processing, cleanliness management, Digital Twin, Nvidia Omniverse Isaac Sim}

\vspace{-20pt}
\section{Introduction}
\label{sec:intro}
\vspace{-10pt}

Nowadays, more and more cities and businesses are turning to advanced solutions to maintain order and cleanliness in public and private spaces. One such innovative tool is video surveillance, sometimes used together with IoT (Internet of Things) devices \cite{sheng2020}. Initially used mainly for security purposes, it is now also increasingly used in cleanliness management \cite{balchandani2017}. CCTV systems, equipped with modern cameras and software, make it possible not only to monitor the state of cleanliness in real time, but also to identify areas in need of intervention and to coordinate the work of cleaning services.
The use of video surveillance in cleaning has a~number of benefits, especially in high-traffic areas. It allows a~quicker response to incidents such as littering in inappropriate places, vandalism or pollution of public spaces. Thanks to image analysis, it is also possible to optimise cleaning schedules and manage human resources more efficiently. Video surveillance is therefore becoming an indispensable element in a~cleanliness management strategy, contributing to improving the quality of life of residents and the aesthetics of the environment.

This paper presents a discussion of selected problems related to the design of a cleaning system using vision processing. The analysis of litter detection, the occupancy of litter bins, the detection of stains on the floor and the determination of the location of the cleaning crew based only on video surveillance data are identified as key areas of focus. The various aspects and practical applications of the digital twin for creating, simulating, training and testing these systems using the example of cleaning maintenance are also discussed, using a~railway station as an example \cite{industryDT}.
We present the advantages and benefits of creating a~digital virtual model of a~public space, using a~railway station as an example, and apply advanced image processing and analysis methods.

Our contribution is the development of a~vision system for maintaining cleanliness in high-traffic areas, using Digital Twin technologies as an example of a~railway station.
The system incorporates key elements to detect and analyse cleanliness situations in real time. 
The system's objective is to check the occupancy of litter bins, monitor cleanliness, and convey data to the cleaning services. The monitoring of the space is conducted in real time, thereby enabling the prompt detection of events. Due to design constraints, the system utilises solely video surveillance data, thus avoiding any interference with the building infrastructure. The anticipated benefits of employing this system are increased cleanliness, reduced cleaning costs (cleaning only when necessary) and maintenance management analysis.
To the authors' knowledge, this is the first vision system of its kind based on Digital Twin technology and the Nvidia Omniverse Isaac Sim.
Section \ref{sec:related_work} presents state-of-the-art methods for detecting litter, bins, people, segmenting stains and analysing their status.
Section \ref{sec:proposed} describes the proposed vision system.
Section \ref{sec:demo} contains the results of the system on real data and Section \ref{sec:conslusion} provides a~summary and future works discussion.

\vspace{-20pt}
\section{Related work}
\label{sec:related_work}
\vspace{-10pt}

Cleanliness in high-traffic areas such as train stations, subways, airports and shopping centres is not only a~matter of aesthetics or traveller convenience, but also an important factor affecting public health and community quality of life. Eliminating potential hazards, such as slips on wet floors or accidents caused by objects left on the ground, is vital to ensure safe travel conditions.
As such, an adequate cleanliness infrastructure, regular cleaning, appropriately placed bins and traveller education on tidiness are key to maintaining high standards of cleanliness in high-traffic areas.

\vspace{-20pt}
\subsection{Vision systems for maintaining cleanliness}
\vspace{-10pt}

One of the solutions that influence the appropriate maintenance of cleanliness is the use of automatic or semi-automatic video surveillance systems. 
Their tasks include: 
\begin{itemize}
   
    \item \textbf{The automated detection of litter or other waste, as well as real time image analysis and identification of objects} such as paper, bottles or other waste, which allows for a~faster response from the cleaning services. Different methods based on neural network architectures are used, such as YOLO \cite{xiao2023, lin2021, bawankule2023, wu2021}, SSD \cite{ma2020}, RCNN \cite{chowdhury2022}, vision transformers \cite{li2023}, solutions combining several network architectures \cite{li2022} or comparing the effectiveness of many different models \cite{patel2021, fulton2019, mandhati2024}, but also classical approaches combined with an SVM classifier \cite{salimi2018}. These models are trained on publicly available datasets, including TACO \cite{proenca2020}, UAVVaste \cite{kraft2021}, TrashBox \cite{kumsetty2022}, pLitterStreet \cite{mandhati2024} and others. However, in the solutions custom datasets are often proposed, tailored for specific, less typical applications -- there is no general dataset. A~more comprehensive analysis of published solutions and datasets can be found in the overview article \cite{abdu2022}.

    \item \textbf{Supporting the management of cleaning staff by optimising their activities} -- Cameras can monitor areas with the highest traffic, indicating places that need more frequent cleaning. This makes it possible to distribute tasks more efficiently among cleaning staff. Classic image processing methods -- correlation filter, optical flow, object tracking networks -- Siamese networks, recurrent neural networks (RNNs), as well as static and probability tools for data collection and prediction can be used for this.

    \item \textbf{Prevention of acts of vandalism and pollution} -- The presence of cameras acts as a~deterrent to individuals who might commit acts of vandalism, theft or deliberate pollution of public spaces. In the event of such situations, camera footage can serve as evidence in investigations and help punish the perpetrators. Many solutions are based on object detection, for example, detecting unwanted human presence in a~restricted area \cite{vandalism1} or detecting and clustering dangerous objects \cite{vandalism2}. For the prevention of vandalism in places such as a~train station, it may be more effective to formulate the problem as anomaly detection in a~video sequence. For this task, deep features that have some kind of time-dimensional information are often used \cite{vandalism3, vandalism4}.

\end{itemize}

\vspace{-20pt}
\subsection{Digital Twin}
\vspace{-10pt}

It should also be noted that the effectiveness of video surveillance depends on the proper installation and configuration of the system. 
By placing cameras in strategic locations, it is possible to ensure high quality data covering the entire analysed space.
An equally important aspect is the proper preparation of algorithms for the detection and analysis of video surveillance data. 
Advanced methods also need adequate evaluation to indicate the reliability of the system in different environmental conditions, as well as to simulate various test scenarios, including rare ones.

One of the best technologies for mapping real objects is the use of a~Digital Twin. 
This is the concept of digital mapping of physical objects, processes or systems.
It allows the mapping, using a~BIM (Building Information Modelling) model, of the geometry, appearance including textures of the real object and the surrounding environment.
Using simulations and appropriate algorithms, it models the actual behaviour of static and dynamic objects, people and the interactions between them.
Furthermore, by using a~Digital Twin, it is possible to fuse data from different sources, obtaining a~comprehensive picture of the real object, even during operation.
Examples of tools to create BIM models include Autodesk Revit, and to generate a~virtual environment the Nvidia Omniverse Isaac Sim simulator can be used. The use of the ROS robotic environment to interact with real objects is also possible.

\vspace{-20pt}
\section{The proposed system}
\label{sec:proposed}
\vspace{-10pt}

We propose a~vision-based system for maintaining cleanliness in high-traffic areas, using the example of a~railway station. 
A~virtual model of the railway station was prepared using the Digital Twin technology and Nvidia Omniverse Isaac Sim simulator. 
It was applied to develop a~litter detector, a~bin occupancy level detector, an algorithm for segmenting surface contamination (stains), and a~human detector (including the cleaning crew) with their movement analysis.

\vspace{-15pt}
\subsection{Digital Twin -- reconstruction of the railway station}
\label{sec:simulation}
\vspace{-10pt}

\begin{figure}[h!]
    \centering
    \subfigure[3D model in Autodesk Revit]{
        \includegraphics[width=0.4\textwidth]{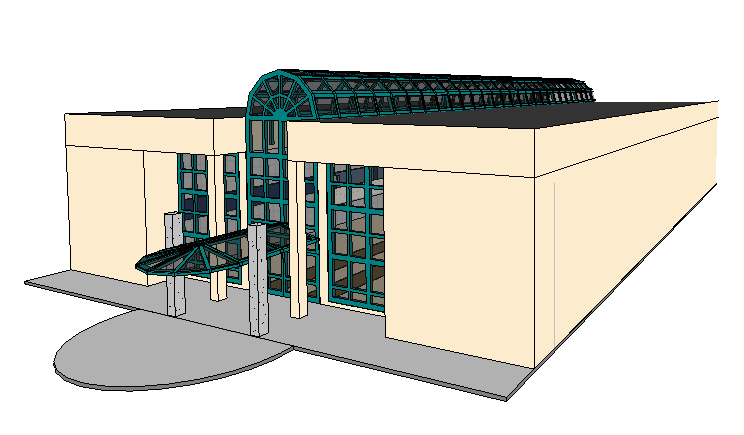}
        \label{fig:image_revit}
    }
    \hfill
    \hspace{-10pt}
    \subfigure[Digital Twin in Nvidia Omniverse Isaac Sim]{
        \includegraphics[width=0.45\textwidth]{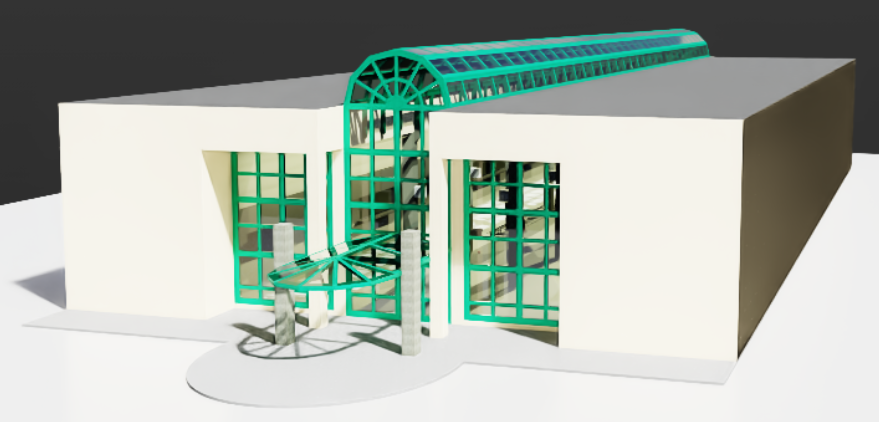}
        \label{fig:image_isaac}
    }

    \vspace{-10pt}
    \caption{3D model of the railway station: \subref{fig:image_revit} designed in Autodesk Revit and \subref{fig:image_isaac} Digital Twin based on the 3D model with added materials and textures.}
    \label{fig:images_revit_isaac}
\end{figure}

The Digital Twin was created using data from Autodesk Revit, which is an advanced Building Information Modelling (BIM) tool that enables creating detailed 3D projects. 
It was used to design an accurate model of the railway station, incorporating all architectural, structural, and installation elements.
The Digital Twin presents a~real model of a~railway station in one of Poland's largest cities.
It was created in one of the best 3D computer simulators -- Nvidia Omniverse Isaac Sim.
Figure \ref{fig:image_revit} shows a~3D model of the railway station made in Autodesk Revit, and Figure \ref{fig:image_isaac} shows the same model in the Nvidia Omniverse Isaac Sim simulator using the appropriate textures and materials.

\vspace{-15pt}
\vspace{-35pt}

%\subsection{Analiza rozmieszczenia kamer}
\subsection{Analysis of camera deployment}
\vspace{-10pt}

\begin{figure}[t]
    \centering
    \includegraphics[width=0.6\textwidth]{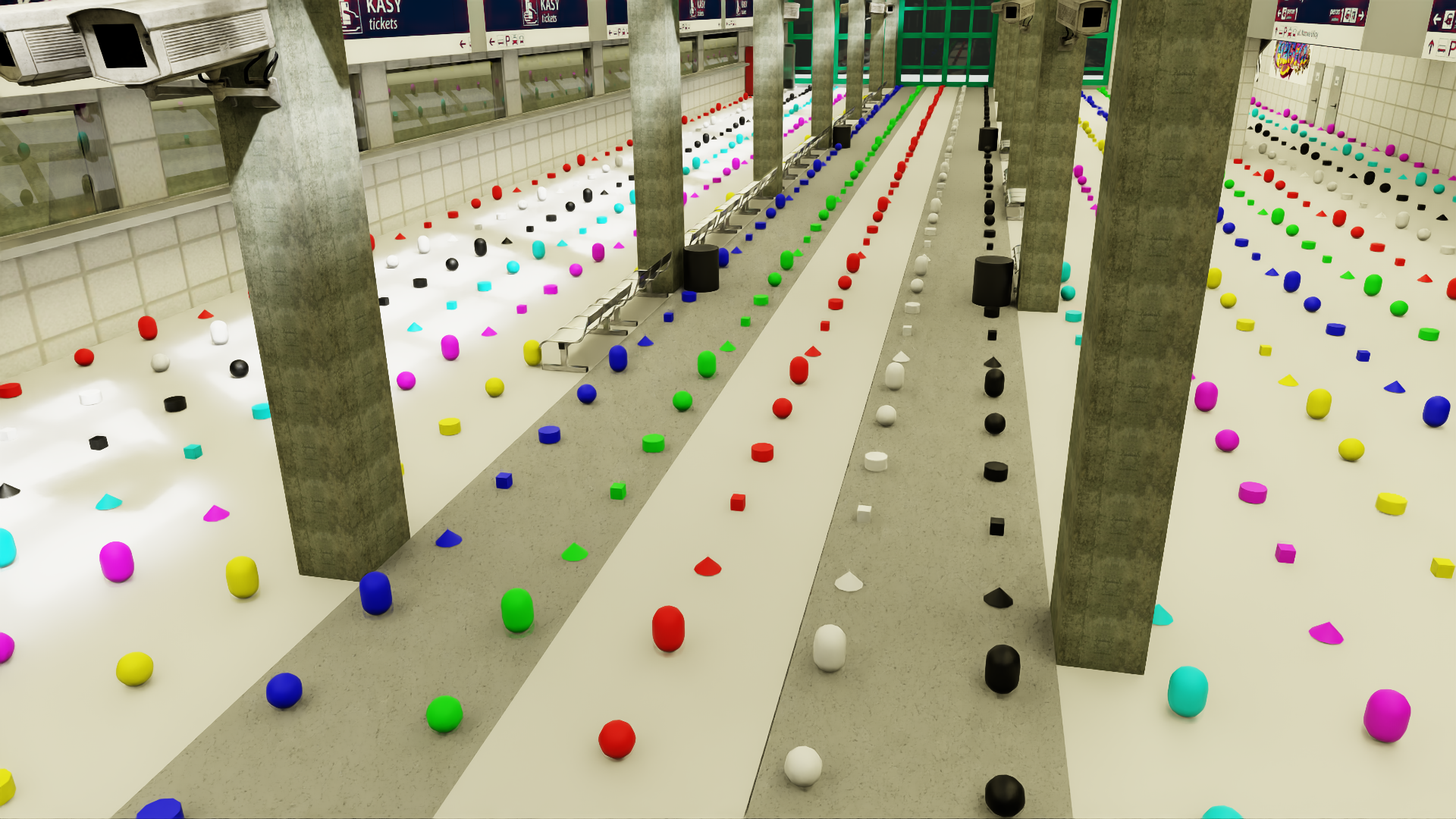}
    %\caption{Wygenerowane proste obiekty 3D, rozmieszczone na regularnej siatce o boku 1 metra, do analizy rozmieszczenia kamer oraz pokrycia terenu dworca przez monitoring wizyjny.}
    \caption{Generated simple 3D objects, arranged on a regular grid, 1 metre apart, to analyse cameras placement and video surveillance coverage of the station area.}
    \vspace{-10pt}
    \label{fig:pokrycie}    
\end{figure}

In order to keep the entire area of the actual station clean, every part of it should be covered by video surveillance.
Therefore, one of the tasks was to propose the placement of a~set of cameras and to analyse the parts of the station covered by their field of view.
During the analysis, more attention was paid to areas with significant passenger flows (entrance, ticket offices, etc.). 
A~script was used to analyse the overall video surveillance coverage of the area, generating simple 3D objects in different colours on a~rectangular grid (for easy distinction), 1 metre apart.
The result was a~set of 1340 objects distributed inside the station, as shown in Figure \ref{fig:pokrycie}.
The visibility of the objects from the individual cameras was then analysed by marking the visible shapes.
In this way, `blind spots' not visible on any of the cameras were determined, which was then used to correct their distribution relative to the original determination.
In further tests, all generated objects were marked -- 14 cameras were sufficient for full coverage.
It can therefore be assumed that in this configuration the entire area of the virtual station is covered by video surveillance.
% Figure \ref{fig:pokrycie_kamer} shows a~top view with the cameras deployed.

\vspace{-15pt}

\subsection{Litter detection}
\vspace{-10pt}

\begin{figure}[!ht]
    \centering
    \includegraphics[width=0.45\textwidth]{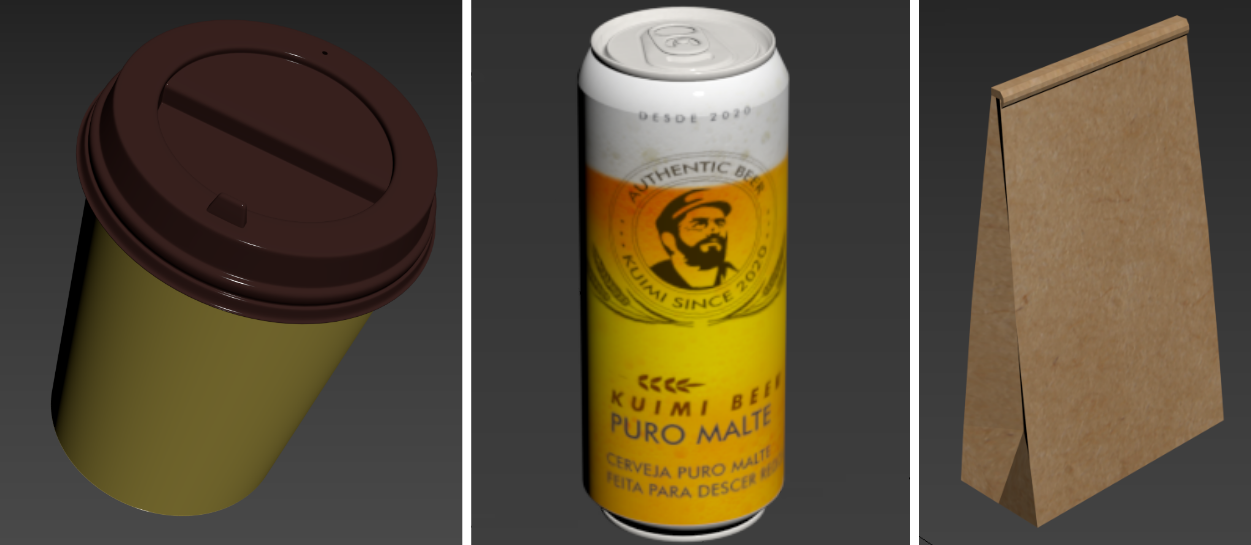}
    %\caption{Przykładowe przygotowane modele śmieci.}
    \caption{Examples of prepared litter models.}
    \vspace{-10pt}
    \label{fig:trash_images}   
\end{figure}

An essential part of the method for detecting litter at the station was the preparation of appropriate 3D models and the their distribution, both loose lying and in bins to determine their degree of occupancy.
Figure \ref{fig:trash_images} shows examples of litter models.

% \vspace{-5pt}
% \vspace{-5pt}

Based on an analysis of several models available on the GitHub platform, the latest version of the popular YOLO (You Only Look Once) network -- YOLOv8 \cite{jocher2023} -- was chosen.
The performance of different versions of this network was tested (in terms of size, starting from the smallest): YOLOv8n, YOLOv8m, YOLOv8x.
As the size of the model increases, the inference time increases, but so does (in the general case) the detection accuracy.
However, these pre-trained models were not suited to litter detection, which required training on a~suitable dataset.
Therefore, several publicly available datasets containing images of different types of litter were selected for training.
The evaluation was then performed on a~set of images from the station model -- the highest performance was observed for the model trained on the TACO dataset \cite{proenca2020}.
It is worth mentioning here that the datasets contained only images of real litter, but due to good knowledge generalisation by the trained model, generated 3D models of litter in the simulation environment were correctly detected.
Some problems arose in the case of litter that was small or far away from the camera, as it was not detected.
Therefore, another issue to consider was the resolution of the cameras used and the modification of the litter detection method.
Different options can be easily tested in the station model, so an identical set of litter was placed multiple times at increasing distances from the camera.

\begin{table*}[t]
%\caption{Porównanie czasów inferencji (w milisekundach) dla różnych rozdzielczości kamer oraz wybranych modeli -- z metodą SAHI (with SAHI) oraz bez niej (no SAHI), dla sieci YOLOv8n oraz YOLOv8x.}
\caption{Comparison of inference times (in milliseconds) for different camera resolutions and selected models -- with SAHI and without SAHI, for YOLOv8n and YOLOv8x networks.}
\centering
\label{tab:inf_times_yolo}
\begin{tabularx}{0.61\textwidth}{lcccc}%{@{} l *{5}{c} c @{}}
\toprule
 & \multicolumn{2}{c}{no SAHI} & \multicolumn{2}{c}{with SAHI} \\ 
Resolution & YOLOv8n & YOLOv8x & YOLOv8n & YOLOv8x \\ 
\midrule
640 $\times$ 480 & 11 & 39 & 704 & 805 \\
720 $\times$ 576 & 11 & 32 & 789 & 1022 \\
1024 $\times$ 768 & 15 & 43 & 1021 & 1230 \\
1280 $\times$ 720 & 11 & 30 & 921 & 1257 \\
1920 $\times$ 1080 & 13 & 38 & 1369 & 2321 \\
3840 $\times$ 1920 & 9 & 36 & 2096 & 4298 \\
\bottomrule
\end{tabularx}
\end{table*}

\begin{figure}[t]
    \centering
    \subfigure[Without SAHI]{
        \includegraphics[width=0.45\textwidth]{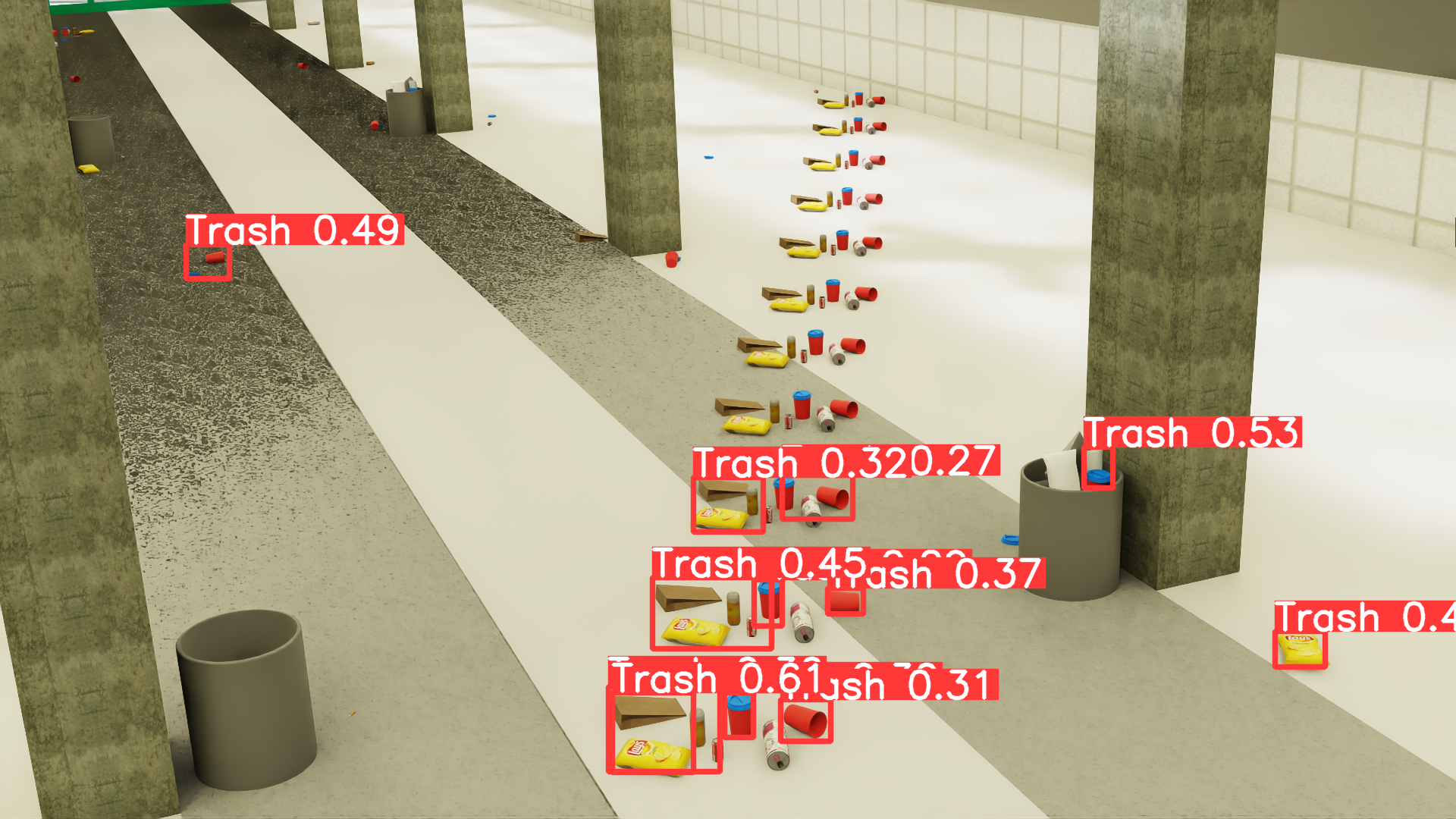}
        \label{fig:detection_no_sahi}
    }
    \hfill
    \subfigure[With SAHI]{
        \includegraphics[width=0.45\textwidth]{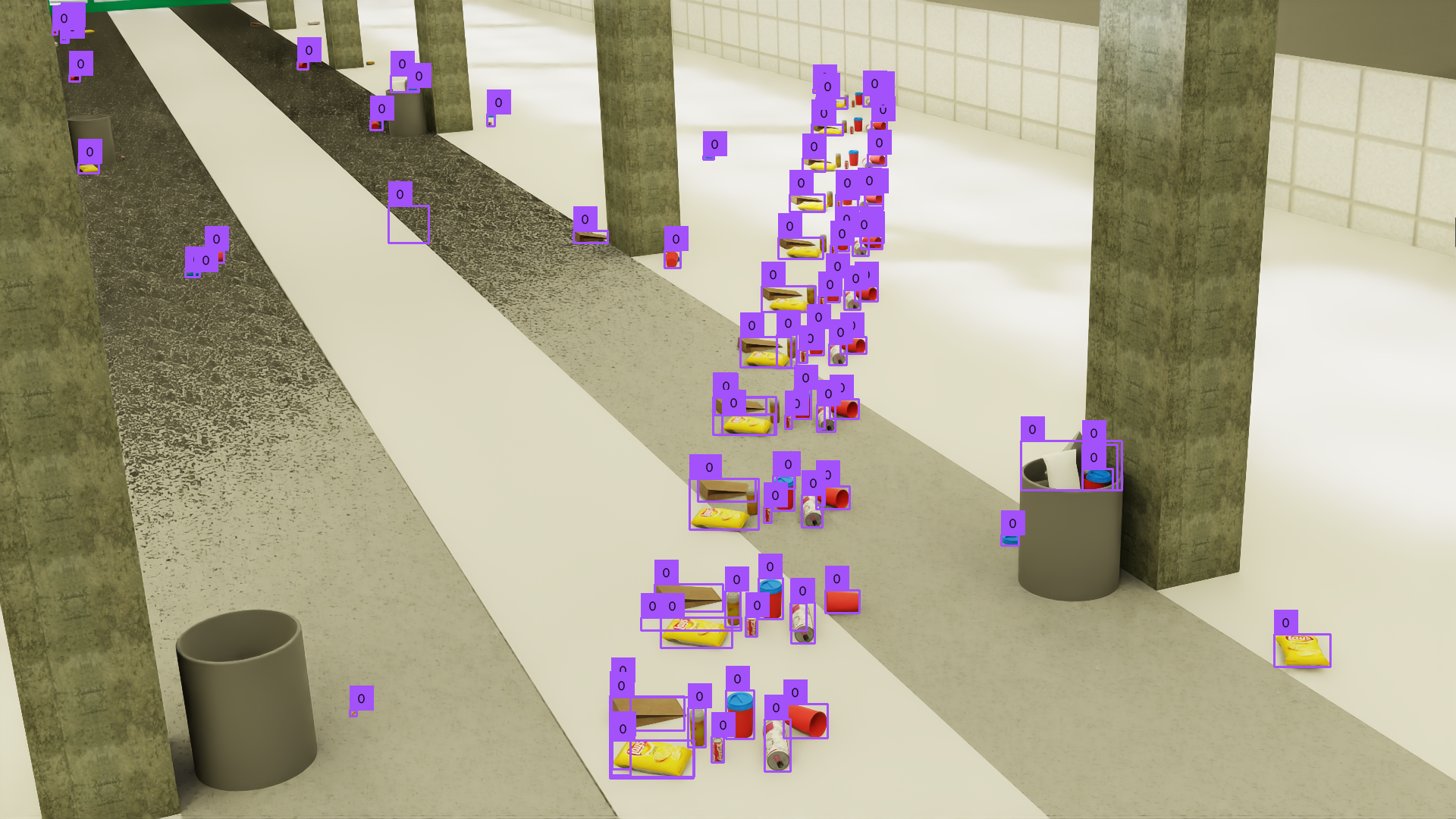}
        \label{fig:detection_sahi}
    }
    %\caption{Porównanie działania nauczonego modelu sieci YOLOv8x z dodatkowym zastosowaniem metody SAHI do detekcji małych obiektów.}
    \vspace{-10pt}
    \caption{Comparison of the performance of the trained YOLOv8x network model with the additional use of the SAHI method for small object detection.}
    \vspace{-10pt}
    \label{fig:sahi_comparison}
\end{figure}

Performance was tested for a~range of resolutions: $640 \times 480$, $720 \times 576$, $1024 \times 768$, HD -- $1280 \times 720$, Full HD -- $1920 \times 1080$ and 4K -- $3840 \times 2160$ pixels.
However, the differences for the higher resolutions were imperceptible due to the scaling of the input image to a~fixed size of $640 \times 360$ pixels.
Therefore, the SAHI (Slicing Aided Hyper Inference) method \cite{akyon2022} was used, which divides the input image into fragments, performs detection on each one, and finally combines the results from all fragments.
The use of this method has significantly improved the performance of the network, especially for higher camera resolutions.
However, at the same time, it significantly increases the inference time up to 5 seconds for a~single 4K frame.
Figure \ref{fig:detection_no_sahi} shows the detection result for a~Full HD image without the SAHI method, while Figure \ref{fig:detection_sahi} the result of the performance after applying this method.

A~comparison of the average inference times for the different network variants with and without the SAHI method, for different camera resolutions, is provided in Table \ref{tab:inf_times_yolo}.
In the case of higher camera resolutions, the running times are quite long, while a~significant benefit is then higher detection accuracy.
In an application of this type, involving the detection of static litter lying in the station area, the operation does not need to be performed in real time -- detecting litter every few seconds is sufficient, while the accuracy and precision of detection should be considered as more important.

% \vspace{-15pt}
% \vspace{-35pt}
\vspace{-20pt}
%\subsection{Detekcja poziomu zajętości kosza na śmieci}
\subsection{Detection of bin occupancy level}
\vspace{-10pt}

One of the specified requirements was that only vision be used to detect bin occupancy To the best of our knowledge, there are no known methods that use only vision to determine bin occupancy. Typically, other types of sensors are used for this task. The bin fullness detection vision algorithm must be adapted to the type of bin to be found in the target environment. Due to the lack of public datasets containing images of full and empty litter bins, it is not possible to use machine learning based approaches. 
% A~time-consuming and costly preparation of a~large dataset containing images of empty and filled bins, from different angles and with different contents, would be necessary. 
It was therefore decided to use a~``classical'' image processing algorithms.

\begin{figure}[h!]
    \centering
    \subfigure[]{
        \includegraphics[width=0.4\textwidth]{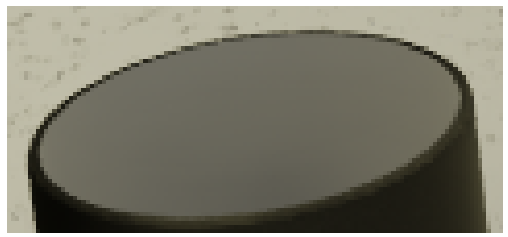}
        \label{fig:KoszPusty}
    }
    \hfill
    \subfigure[]{
        \includegraphics[width=0.43\textwidth]{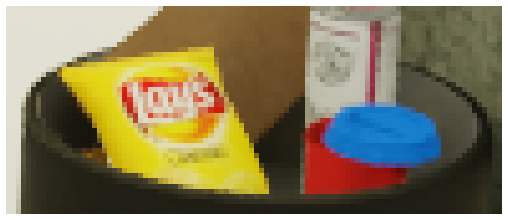}
        \label{fig:KoszPelny}
    }
    %\caption{Porównanie widoku z tej samej kamery na pusty i pełny kosz na śmieci.}
    \vspace{-10pt}
    \caption{Comparison of the view from the same camera on \subref{fig:KoszPusty} empty and \subref{fig:KoszPelny} full bin.}
    \vspace{-10pt}
    \label{fig:Kosze}
\end{figure}

It was assumed that the analysed bin is an opaque cylinder that is not closed from the top. It is therefore necessary to assess its filling based on the contents seen through the top opening. It is also assumed that the ROI around the open part of the dustbin is given. Figure \ref{fig:Kosze} shows the camera image for an empty and a~full bin.

\vspace{-10pt}

\begin{figure}[!t]
    \centering
    \includegraphics[width=0.5\textwidth]{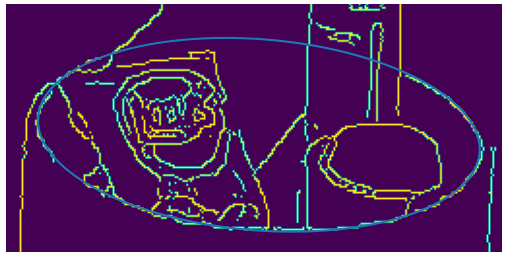}
    %\caption{Wykryte krawędzie oraz wynik detekcji elips za pomocą transformaty Hougha.}
    \caption{Detected edges and the result of ellipse detection using the Hough transform.}
    \vspace{-10pt}
    \label{fig:Hough}
\end{figure}

A~custom algorithm has been developed to determine the occupancy of a~bin, based on an analysis of the view of the edges of the bin and its interior.
Initially, a~colour space conversion to HSV (Hue, Saturation, Value) space is performed. Based on the results, the S-component was found to give the best results for determining the top edge of the bin using Canny's algorithm. Therefore, this component was further used, but it was first subjected to Gaussian filtering. A~Hough transform was performed on the image containing the edges in order to detect the ellipse, which is the top edge of the analysed bin. 
Figure \ref{fig:Hough} shows the detected edges and highlights the ellipse in blue.

The final stage of bin fullness detection was divided into two parts. In the first, the pixels inside the ellipse were marked using morphological operations and then the standard deviation of these pixels was calculated for the processed component S. If it was greater than a~given threshold, it was assumed that the inside of the bin was inhomogeneous and therefore litter was visible. A~decision was therefore made that the bin was full.
In the second part, it was assumed that, for a~full bin, some of the litter may protrude above the top edge of the bin. As a~result, the further edge may not be fully visible in the camera image. An additional image was therefore created in which only the top edge of the ellipse was marked. It was further thickened using morphological dilation by 1 pixel. The edge pixels were then counted (after Canny's algorithm), which are in the area marked as the top edge of the ellipse. Based on the determined ellipse parameters, the number of edge pixels that should be visible in the case of an empty bin was also estimated.
If the number of visible edge pixels was less than a~given fraction of the estimated value, it was assumed that some litter was obscuring the top edge, so the bin was full.

%\subsection{Detekcja plam}
\vspace{-20pt}
\subsection{Stain detection}
\vspace{-10pt}

Stain detection is a~relatively broad issue due to the different sources of contamination.
These translate into different textures and colours of the areas to be segmented in the camera image.
Therefore, the work described here focuses on the detection of water stains (puddles), which represent one of the most common cases of surface contamination in public spaces.
Water stains are usually distinguished by a~local increase in the reflectance of the analysed surface while lacking a~characteristic colour.
For this reason, when developing the segmentation algorithm, it was decided to reuse the processing of the video stream in HSV colour space.
In this model, the saturation component manifests the aforementioned reflectance disorder.
This effect was decided to be further enhanced by using the CLAHE (Contrast Limited Adaptive Histogram Equalisation) algorithm on the perceptual lightness component in the CIELAB colour space as an intermediate step in the conversion from RGB to HSV.
On such a~prepared image, different segmentation possibilities were analysed using both global and adaptive thresholding.
In the end, it was decided to use the former approach, with the threshold \textit{thr\_S} as the hyperparameter of the algorithm.
The segmentation process was further complemented by dilation, median filtering and connected component analysis (CCA), which allows the removal of unwanted small-scale artefacts.
The results obtained were also filtered over time.
A~given pixel was only labelled as a~background element when it had not been segmented as a~blob by consecutive \textit{thr\_history} frames (another hyperparameter of the algorithm).
In this way, a~relatively efficient and computationally effective stain segmentation algorithm was obtained.
Its performance can be adapted to specific conditions by changing two hyperparameters: \textit{thr\_S} and \textit{thr\_history}.
Their proper selection should be part of the system calibration process when it is run on a~real object.

%\subsection{Detekcja ludzi i analiza ruchu}
\vspace{-20pt}
\subsection{Human detection and motion analysis}
\vspace{-10pt}

\begin{figure}[t]
    \centering
    \subfigure{
        \includegraphics[width=0.45\textwidth]{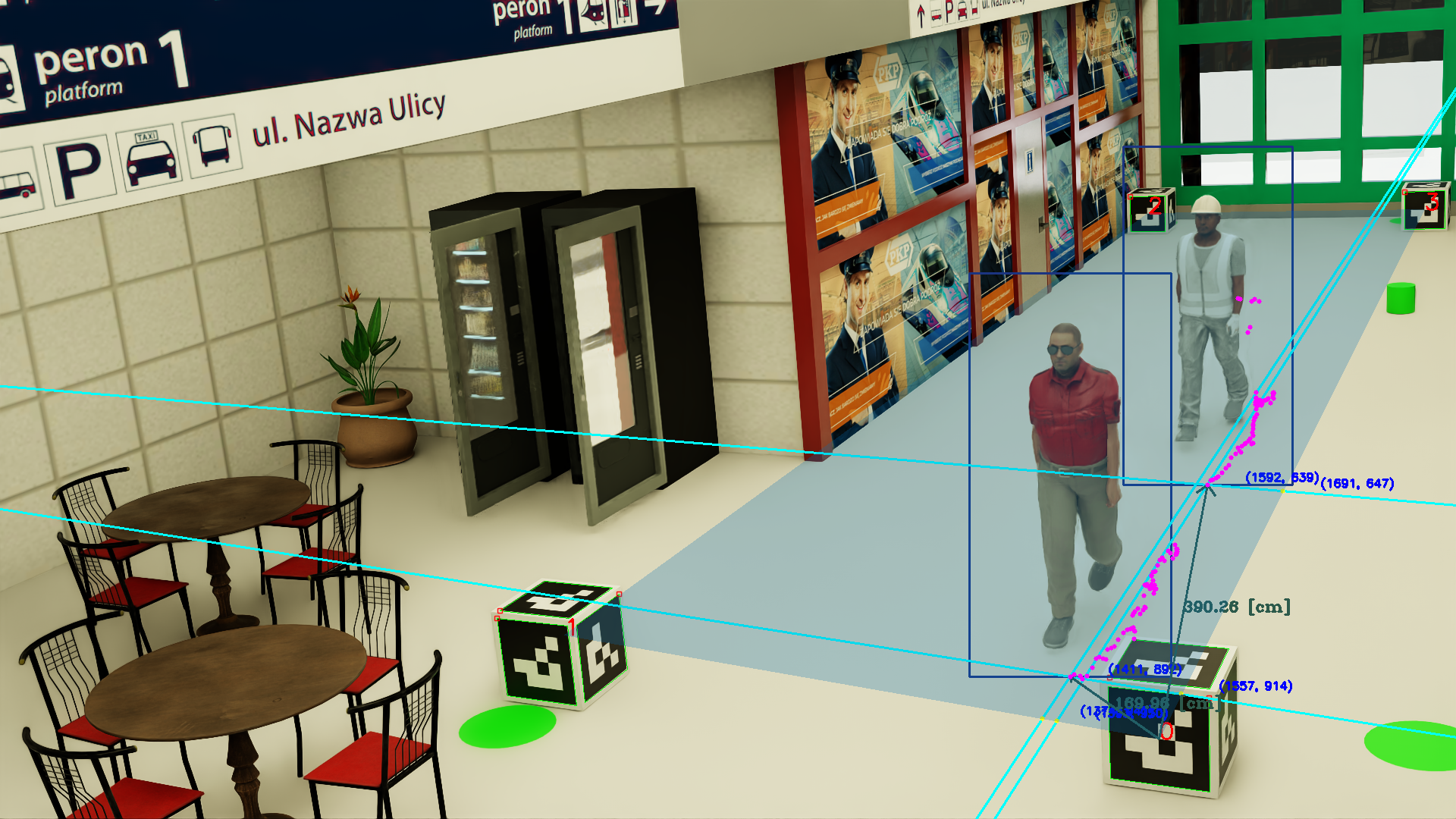}
        \label{fig:image1}
    }
    \hfill
    \subfigure{
        \includegraphics[width=0.45\textwidth]{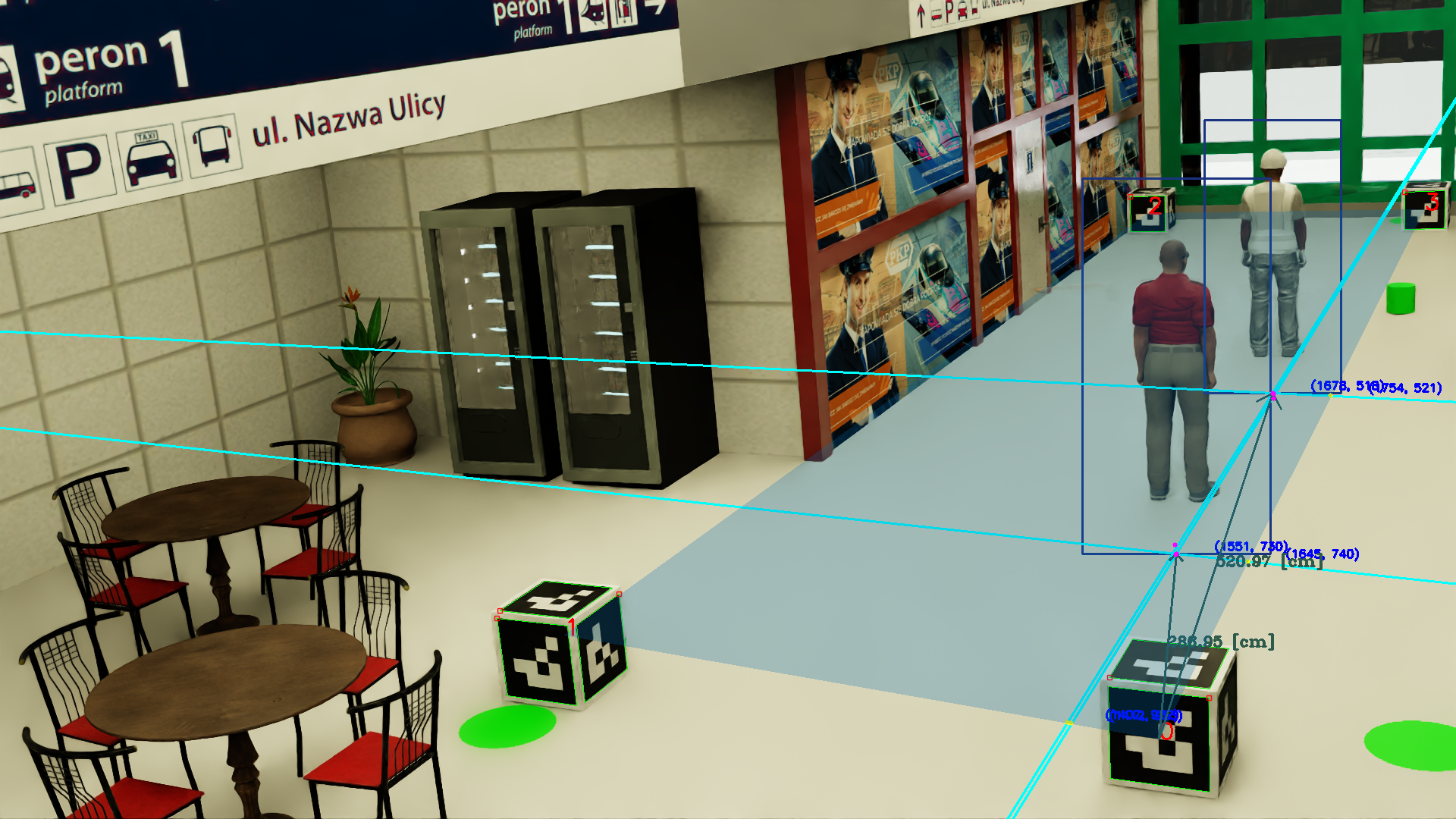}
        \label{fig:image2}
    }
    %\caption{Przykładowy wynik detektora ludzi wraz z analizą ruchu.}
    \caption{Example human detector result with motion analysis.}
    \vspace{-10pt}
    \label{fig:det_people}
\end{figure}

Analysis of the areas that need cleaning and the location of cleaning staff and equipment requires the correct calibration of the entire camera system. This can be done using the video surveillance data available at the considered railway station. 
Therefore, a~special algorithm has been developed to perform the calibration under these conditions.
The process begins with the delineation of the analysis area for pedestrian and other objects using cubes -- an ArUco marker is placed on each wall. 
With these markers, their precise location in space and identification is possible. On the basis of the collected data, the corresponding perspective and vanishing points are determined, allowing the position of other objects to be accurately mapped. 
For the detection of people, the classic HOG+SVM method was used, but neural networks, such as the YOLO family, can also be applied. Based on simple geometrical relationships, the position of objects can be determined relative to the selected cube or relative to other detected objects. In addition, the system records the position history of all detected objects.
Figure \ref{fig:det_people} shows the detection of people in a~designated area, followed by the determination of the actual distance (description in green in centimetres) from the cube closest to the camera (bottom right corner of the image).
It is also easy to transfer the detected positions to a~2D map of the object (top view).

% \vspace{-5pt}

\vspace{-20pt}

\section{Real-world demonstrator}
\label{sec:demo}
\vspace{-10pt}

\begin{figure}[t]
    \centering
    \subfigure{
        \includegraphics[width=0.45\textwidth]{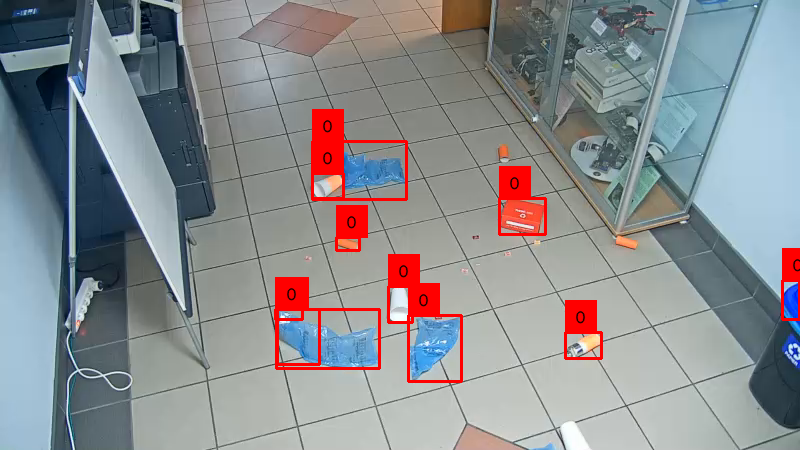}
        \label{fig:demo_trash1}
    }
    \hfill
    \subfigure{
        \includegraphics[width=0.45\textwidth]{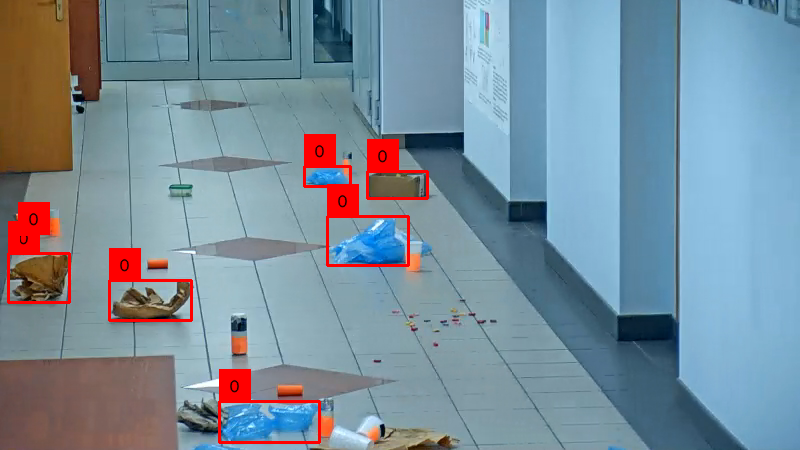}
        \label{fig:demo_trash2}
    }
    %\caption{Przykładowe wyniki detekcji śmieci z różnych ujęć kamery na uczelnianym korytarzu.}
    \caption{Example results of litter detection from different camera shots in a~university corridor.}
    \vspace{-15pt}
    \label{fig:demo_trash}
\end{figure}

To test the developed ways of maintaining cleanliness in real world, we prepared a~simple demonstrator in the corridor of one of the university buildings.
The plan was to test the algorithms in the real world and to investigate the sim-to-real gap phenomenon of differences between the performance in simulation and reality.
A~camera with variable orientation and field of view was used to test the performance of the algorithms from simulated different sensors.
As part of the tests, different types of litter were scattered in the corridor.
The scene was then recorded from several perspectives, simulating different camera parameters.
It was necessary to retrain the YOLOv8 network model on another dataset (Open Image V7) in order to better adapt the performance to the real sequences.
Example results are provided in Figure \ref{fig:demo_trash}.
In the general case, litter is detected correctly, although there are some isolated false-positive detections when other objects not present in the simulated station (glass display cases, posters, etc.) are considered to be litter -- but some of these errors are caused by the sheer differences between the simulated space (the station) and the demonstrator (the university corridor).

% \vspace{-10pt}

\vspace{-20pt}
\section{Conclusion}
\label{sec:conslusion}
\vspace{-10pt}

Our work presents a~vision system designed to support cleaning in high-traffic areas, using Digital Twin technology on the example of a~railway station.
A~digital model of the train station was created from a~BIM model in Nvidia Omniverse Isaac Sim simulator. The vision system includes a~litter detector, a~bin occupancy detector, a~surface contamination (stain) segmentation algorithm and a~human detector (including the cleaning crew) with the ability to analyse their movement.
By using the simulator, it was possible to prepare different test scenarios and tune the parameters of the algorithms. In the case of the litter detector, tests were carried out to assess the impact of resolution on detection accuracy and efficiency, as well as to evaluate the ability to correctly identify small items using a~YOLOv8 neural network. For the bin occupancy algorithm, a~fast classical analysis method was proposed. Classical segmentation with colour space and shape analysis was used for stain detection. Classical methods and simple geometrical relationships were also used to detect people and determine their positions in the railway station. In this way, it was possible to analyse the way of movement of a~person and its distance from other objects.

At this preliminary stage, the proposed system is not yet connected to the physical object. However, it is anticipated that, in the final stage, the data flow will occur on an ongoing basis.
In the future works, we also plan to extend our system to include a~detector that signals acts of vandalism and the littering in inappropriate places. In addition, we intend to improve the proposed algorithms, develop occupancy assessment mechanisms for other types of bins, carry out an analysis of the detection of different types of stains and dirt that occur in hard-to-reach places such as wall corners, near bins or benches at the railway station. The next step is to analyse the sim-to-real gap issue using real video sequences from CCTV cameras from the station. 
Quantitative and qualitative tests, including system performance analysis, will be carried out to determine the requirements that the video system should meet in terms of frequency of operation.

\begin{acknowledgement}
Funded by the European Union. Views and opinion expressed are however those of the author(s) only and do not necessarily reflect those of the European Union. Neither the European Union nor the granting can be held responsible for them. This project has received funding from the European Union's Horizon Europe research and innovation programme under Grant Agreement No 101101966.
\end{acknowledgement}

\vspace{-20pt}

\end{document}